\documentclass[5p,times,sort&compress]{elsarticle}

\makeatletter
\def\ps@pprintTitle{%
  \let\@oddhead\@empty
  \let\@evenhead\@empty
  \let\@oddfoot\@empty
  \let\@evenfoot\@oddfoot
}
\makeatother

\usepackage{booktabs}
\usepackage{enumitem}
\usepackage{multirow}
\usepackage{amsmath}
\usepackage{stfloats}
\usepackage{pifont}
\usepackage[hidelinks,colorlinks=true, urlcolor=blue,]{hyperref}

\newcommand{\cmark}{\ding{51}}%
\newcommand{\xmark}{\ding{55}}%

\journal{Neurocomputing}

\begin{document} \sloppy

\begin{frontmatter}

\title{RAPID: Enabling Fast Online Policy Learning in\\Dynamic Public Cloud Environments}

\author[label1]{Drew Penney}
\author[label2]{Bin Li}
\author[label1]{Lizhong Chen\corref{cor1}}
\author[label2]{Jaroslaw J. Sydir}
\author[label2]{Anna Drewek-Ossowicka}
\author[label2]{\\Ramesh Illikkal}
\author[label2]{Charlie Tai}
\author[label2]{Ravi Iyer}
\author[label2]{Andrew Herdrich}
\address[label1]{School of Electrical Engineering and Computer Science, Oregon State University}
\address[label2]{Intel Corporation}

\cortext[cor1]{Corresponding author. Email address: chenliz@oregonstate.edu}

\begin{abstract}
Resource sharing between multiple workloads has become a prominent practice among cloud service providers, motivated by demand for improved resource utilization and reduced cost of ownership. Effective resource sharing, however, remains an open challenge due to the adverse effects that resource contention can have on high-priority, user-facing workloads with strict Quality of Service (QoS) requirements. Although recent approaches have demonstrated promising results, those works remain largely impractical in public cloud environments since workloads are not known in advance and may only run for a brief period, thus prohibiting offline learning and significantly hindering online learning. In this paper, we propose RAPID, a novel framework for fast, fully-online resource allocation policy learning in highly dynamic operating environments. RAPID leverages lightweight QoS predictions, enabled by domain-knowledge-inspired techniques for sample efficiency and bias reduction, to decouple control from conventional feedback sources and guide policy learning at a rate orders of magnitude faster than prior work. Evaluation on a real-world server platform with representative cloud workloads confirms that RAPID can learn stable resource allocation policies in minutes, as compared with hours in prior state-of-the-art, while improving QoS by 9.0x \textit{and} increasing best-effort workload performance by 19-43\%.
\end{abstract}


\end{frontmatter}



\section{Introduction}

On-demand workload deployment in large-scale data centers, supported by the cloud computing paradigm, has evolved as a highly scalable solution for growing compute demand. As part of this evolution, cloud service providers have continued to explore methods to improve server resource utilization and thereby reduce total cost of ownership. In particular, workload co-scheduling has become a prominent practice in which multiple workloads are concurrently executed on the same physical server with shared resources. Contention for these shared resources can, however, introduce significant transient degradation in workload performance and compromise Quality of Service (QoS). Mitigating these fine-grained resource contention behaviors has proven difficult with cluster-level task scheduling alone, which generally cannot achieve higher than 50-60\% average CPU utilization without degrading QoS \cite{alibaba_trace_analysis, borg_2020}. Further increases in workload diversity stemming from growing cloud service adoption \cite{flexera2022} will likely exacerbate these limitations, thus highlighting the need for more fine-grained resource management.

Ongoing research has sought to enable more efficient co-scheduling by carefully managing the resource allocation for each workload, thereby reducing resource contention. Ideally, high-priority (HP) workloads with strict QoS goals should be allocated sufficient resources to accommodate current demand while all remaining resources should be allocated to best-effort (BE) workloads (i.e., those without strict performance requirements). Various methods to perform this dynamic resource allocation have been proposed, with promising results in diverse scenarios, yet these methods remain largely impractical in public cloud environments due to limited workload information and widely varying resource demands. Specifically, in a public cloud, we generally do not have prior knowledge of workload characteristics so resource allocation policies must be adapted online; controllers that do not support online policy adaptation, such as those based on heuristics and control theory, may therefore fail to provide acceptable QoS. Cloud workloads may also only be run for a brief period of time, typically less than 1-2 hours \cite{azure2017}. Meanwhile, initial policy learning can take several hours for the current state-of-the-art controllers based on deep reinforcement learning \cite{Twig2020, firm2020, PROMPT2022}, during which substantial QoS degradation can occur. Moreover, existing machine-learning-based methods, although technically adaptive, often adapt far too slowly to provide practical benefits. Overcoming these limitations necessitates a new approach that can quickly adapt to new operating conditions without sacrificing QoS guarantees.

In this paper, we propose \textbf{R}esource-\textbf{A}ware \textbf{P}redictions for \textbf{I}mmediate Allocation \textbf{D}ecisions (RAPID), a novel framework for fast and fully-online resource allocation policy learning in public cloud environments. RAPID bridges the gap between fine-grained resource contention behaviors and coarse-grained QoS guarantees using a domain-knowledge-inspired QoS prediction mechanism and an efficient deep reinforcement learning controller. This QoS predictor synthesizes information from multiple timescales and, even with limited online samples, is shown to provide effective feedback on resource allocation policy decisions at a rate orders of magnitude faster than is possible in prior work. This near-instantaneous feedback, in turn, enables RAPID to completely decouple resource allocation policy decisions from direct QoS measurements and achieve stable policies in minutes, rather than hours as in prior work. Evaluation on a real-world server platform with representative cloud workloads confirms significant practical benefits for both overall resource utilization and QoS. Specifically, RAPID achieves between 19-43\% average improvement in BE workload performance compared with the prior state-of-the-art based on control theory and reinforcement learning while offering much stronger QoS in highly dynamic operating environments.

\section{Related Work}

Resource allocation optimization has become a highly active research area across numerous domains, ranging from wireless communication networks to smart electric grids. In particular, research on resource allocation in cloud computing applications has rapidly evolved along with the demands of cloud service providers.

\textbf{Cluster-level Schedulers:}
One category of work focuses on cluster-level schedulers that assign workloads to compute nodes \cite{quincy2009, delayschedule2010, qclouds2010, bubbleup2011, paragon2013, omega2013, wharemap2013, quasar2014, towards2014, smite2014, multiobjectiveplace2015, tarcil2015, borg2015, qaware2015, hcloud2016, deeprm2016, improving_spark_2017, hierarchical2017, wei2018schedule, autopilot2020, qosfog2021, proactive_alloc_2021, gosh, cosco}. These schedulers are designed to be workload-agnostic, so generally use coarse-grained estimates of per-node resource utilization and per-workload resource utilization, without considering more fine-grained fluctuations in demand. Co-scheduling decisions must therefore be fairly conservative, which limits average per-node resource utilization to around 50-60\% in various industry settings \cite{alibaba_trace_analysis, borg_2020}. Consequently, there is significant opportunity for further, node-level, resource allocation optimization based on these transient resource demands.

\textbf{Limited-Application Node-Level Controllers:}
Approaches for node-level resource management vary dramatically based on hardware capabilities, assumptions, and goals. In particular, early works were limited by available resource partitioning mechanisms, so sought to mitigate resource contention by temporarily pausing BE workload execution \cite{bubbleflux2013, cpi2_2013, maximizing2013, rubik2015}. Various works have also explored simplified co-scheduling scenarios with BE workloads only and therefore optimize aggregate workload performance, relative workload performance (i.e., fairness) or some combination \cite{vconf2009, url2012, ginseng2016, copart2019, alita_2020, satori2021, DRLPart2021}. These methods cannot be used when co-scheduling HP workloads with strict QoS goals, as addressed in our work.

\textbf{General-Purpose Node-Level Controllers:}
More general methods to co-schedule both HP and BE workloads can be split into two categories: those based on conventional methods and those based on machine learning. Among conventional methods, most are based on search, heuristics, and control theory \cite{LLCPartition2013, octopus2015, Heracles2015, Dirigent2016, dcat2018, Parties2018, Adaptive_QoS_Pred_2019, CLITE2020, rambo2021, LIBRA2021}. These methods offer reliable control in static operating environments, but exhibit distinct limitations in more dynamic scenarios. In particular, continuously changing resource demands may require search-based methods to perform frequent and potentially dangerous exploration, which can introduce orders of magnitude degradation in QoS \cite{PROMPT2022}. Similarly, rigid QoS thresholds in heuristic- and control-theory-based methods can lead to sub-optimal, oscillatory control behavior \cite{Hipster2017} while still requiring long measurement intervals (Section \ref{sec:challenges}), thus limiting their ability to mitigate fine-grained (i.e., sub-second) resource contention. These issues can be partially avoided through the use of QoS proxies (e.g., executed instructions \cite{Dirigent2016, LIBRA2021} or cache miss rate \cite{LLCPartition2013, dcat2018}), but thresholds for these proxies may not reliably hold when workload demand varies greatly (Section \ref{sec:evaluation}). Among these works, LIBRA \cite{LIBRA2021} represents the state-of-the-art as a fast and consistent control approach, but lacks support for online adaptation.

Recent work has primarily focused on more versatile machine-learning-based approaches, which promise robust control behavior through data-driven insight and continuous adaptation to new operating environments \cite{Hipster2017, seer2019, RL_Prediction_2020, firm2020, RLDRM2020, Twig2020, sinan2021, DRLPart2021, chen2022alloc, PROMPT2022, OLPart2023}. Nevertheless, comparatively high data requirements can lead to slow training times on the order of hours \cite{Twig2020, firm2020, chen2022alloc, PROMPT2022}. These training times may be tolerable in other applications, for which workloads are known in advance or are run for days/weeks, but become strictly impractical in public clouds as workloads are generally not known in advance and have an average runtime of less than 1-2 hours \cite{azure2017}. As such, any potential benefits are greatly offset by degradation in QoS incurred during initial policy training. These limitations remain even in the state-of-the-art method with a more efficient network architecture, Twig \cite{Twig2020} and Twig+ \cite{PROMPT2022}, which we demonstrate in evaluation. Concurrent work, OLPart \cite{OLPart2023}, proposes an online learning approach based on multi-armed bandits. OLPart is shown to improve scalability when co-scheduling multiple HP and BE workloads but exhibits degraded performance when the control interval is reduced below 2 seconds due to its fundamental dependence upon QoS measurements \cite{OLPart2023}, thus hindering adaptation to new workload behaviors. In contrast, our proposed work is shown to effectively allocate resources at an order-of-magnitude faster control interval and, based on overhead analysis, a millisecond-level control interval in an embedded system implementation. Our work, RAPID, is therefore proposed as the only work to offer both fast and adaptive resource allocation in highly-dynamic public cloud environments.

\section{Preliminaries}

\subsection{Problem Formulation}
Physical compute nodes generally have multiple resources, $R=\{R_1,...,R_n\}$, each with a finite number of discrete units to be allocated, denoted as $R^*=\{R_1^*,..., R_n^*\}$. In practice, there exist dependencies between these resources such that strictly allocating only a subset of these resources, $\hat{R}=\{R_1,...,R_m\}$ ($m\leq n$), can mitigate contention on other resources. As an example, restricting the rate at which a workload accesses offcore data (i.e., memory bandwidth) can reduce its ability to contend for shared cache that would be used as a temporary storage for that data.\footnote{As with most prior work, we assume that CPU cores are not shared between workloads since no existing mechanism can prevent inter-thread contention in private, per-core resources.} Among workloads to be co-scheduled, we consider two priority levels: 1) HP workloads with strict performance (i.e., QoS) targets and 2) BE workloads without strict performance targets. We denote currently co-scheduled workloads as $W=\{W_1^{HP},...,W_x^{HP}, W_1^{BE},...,W_y^{BE}\}$ and HP workload QoS targets as $Q^{HP}=\{Q_1^{HP},...,Q_x^{HP}\}$.\footnote{The set of co-scheduled workloads on a compute node often varies over time as new jobs are started and old jobs are completed. We simplify discussion by focusing on a time window with fixed workloads, although overall methodology remains applicable to dynamic workload sets.} Measured workload performance for workload $W_{w}$, which also varies over time $t$, is then denoted as $P_{t}^{\,meas\,}(W_{w})$. We similarly denote predicted workload performance (discussed in Section \ref{sec:rapid}) for workload $W_{w}$ at time $t$ as $P_{t}^{\,pred\,}(W_{w})$. At each timestep $t$, each workload is allocated a portion of each resource, denoted as $\hat{R}_{r,w,t}$. The goal of workload co-scheduling, as given by Equation \ref{eq:coscheduling}, is to allocate the finite system resources in a manner that maximizes performance for BE workloads while meeting the performance target for all HP workloads.

\begin{equation}
    \begin{aligned}
    \text{maximize}   \qquad & \sum_{w=1}^y P_{t}^{\,meas\,}(W_{w}^{BE})                        \\
    \text{subject to} \qquad & P_{t}^{\,meas\,}(W_{w}^{HP}) \geq Q_w^{HP}   &&   \forall w=1,x  \ \ \& \ \ \forall t=1,T \\
                  \qquad & R_r^* \geq \sum_{w\in W} \hat{R}_{r,w,t}   &&   \forall r=1,m \ \ \& \ \ \forall t=1,T  \\
    \end{aligned}
    \label{eq:coscheduling}
\end{equation}
Note that the resources required to satisfy HP QoS targets can vary dramatically over time along with changes in workload behaviors. Consequently, there is no single, optimal solution. Achieving these objectives requires a dynamic approach that safely minimizes the resource allocation for each HP workload based on current demand, thereby maximizing available resources for BE workloads. Our proposed framework decomposes this task using a separate controller for each HP workload. Discussion in this paper therefore focuses on controller-level design for individual HP workloads, although overall methodology remains applicable to environments with multiple HP workloads (Section \ref{sec:discussion}).

\begin{figure*}[t]
    \centering
    \includegraphics[width=0.97\linewidth]{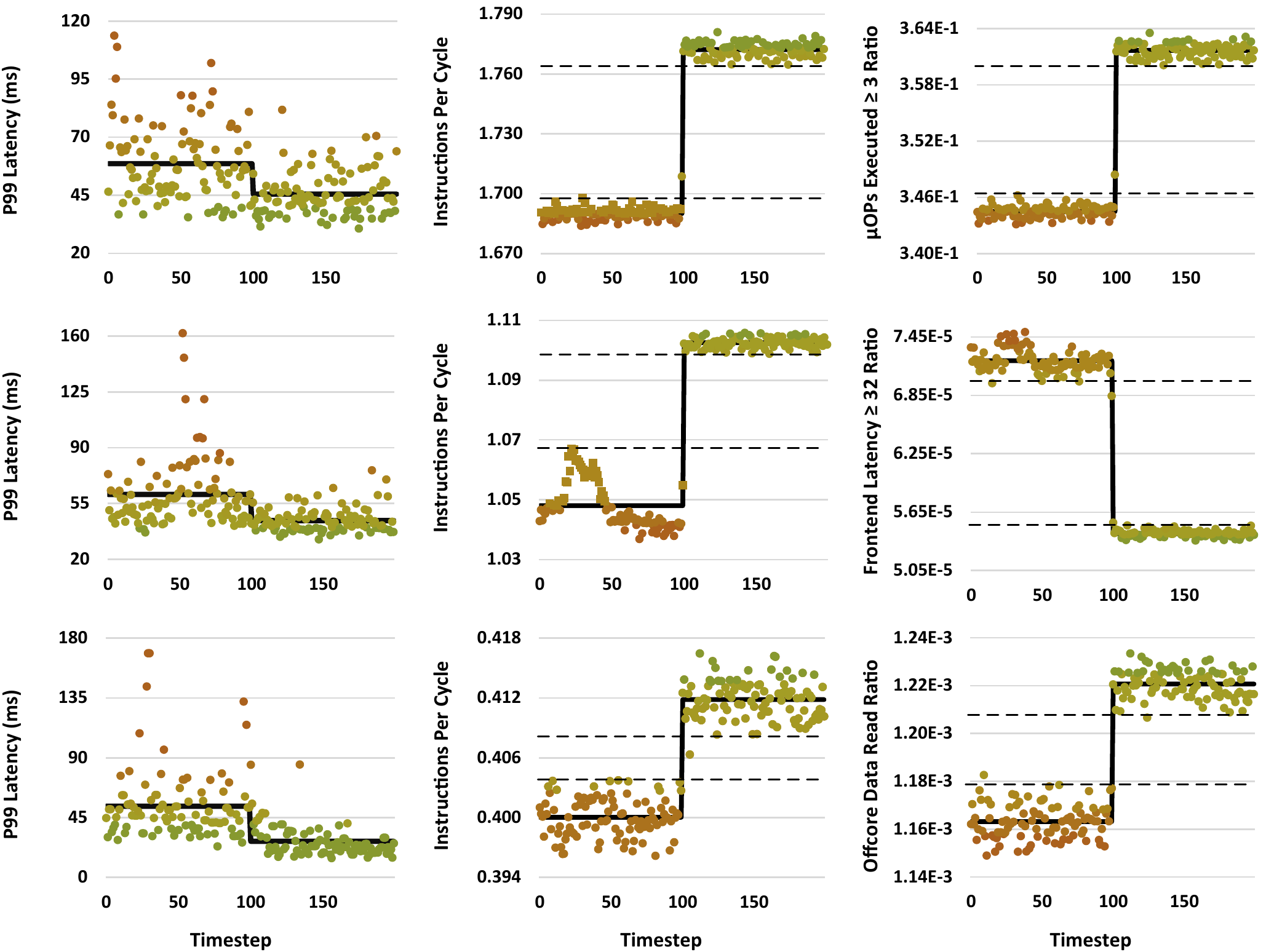}
    \caption{Consistency of various measurement sources when co-scheduling HP and BE workloads. Top row: HP image classification workload (ResNet). Middle row: HP recommendation workload (DLRM). Bottom row: HP web server workload (Nginx). Dots illustrate individual readings at a one second interval while the solid black line illustrates the long-term average before and after BE resource throttling is applied (timestep 100).}
    \label{fig:coscheduling_behaviors}
\end{figure*}

\subsection{Challenges \& Insights} \label{sec:challenges}

Reduction of training time by the required 1-2 orders-of-magnitude is a challenging task due to transient resource demands and transient system response characteristics. These factors introduce substantial uncertainty in policy feedback, which can mislead resource allocation decisions and degrade QoS.

\subsubsection{Uncertainty in QoS}
Variations in resource demand can occur for many reasons, ranging from coarse-grained shifts in workload demand (e.g., requests per second) to fine-grained phases in workload execution. These variations can be quite significant; as highlighted by the Microsoft Azure 2017 dataset, less than 5\% of workloads have an average CPU utilization above 50\% yet nearly 40\% of workloads exhibit spikes in CPU utilization higher than 80\% \cite{azure2017}. These peaks in resource demand can substantially degrade QoS when co-scheduled with other workloads, especially as service providers adopt more aggressive policies with node over-subscription, since overlapping peaks may saturate physical resources \cite{borg_2020}. 

We conducted several experiments to verify these behaviors, as illustrated in Figure \ref{fig:coscheduling_behaviors}. Each experiment (row) co-scheduled a representative HP cloud workload alongside a compute-intensive BE workload. HP workload demand was held constant using an external traffic generator. Latency for each request was measured and then QoS is reported as the latency value higher than 99\% of measurements (i.e., 99th percentile or p99). Experiments began with a period of moderate resource contention (timesteps 0-100), after which BE resource throttling was increased, thereby reducing resource contention (timesteps 100-200). We first highlight that QoS measurements (left column) at a one-second interval can vary by more than 100\% compared to the long-term average, particularly when resource contention is high. As such, individual measurements may fail to appropriately represent long-term workload performance and therefore misguide control policy learning. We further observe that the distribution of QoS measurements before and after resource throttling can remain similar, making it difficult to immediately determine whether a change in resource allocation was actually beneficial. Taken together, it may be necessary to average QoS measurements over a longer period (e.g., 15 seconds \cite{Heracles2015}) in order to obtain low-variance QoS estimates, thereby rendering most existing approaches impractical in highly-dynamic, online learning scenarios.

Resource allocation decisions can instead be made using alternative information sources (e.g., hardware performance counters) as proxies for QoS, but their effectiveness is limited in existing works. In particular, instructions-per-cycle (IPC) is commonly used as the sole information source in non-machine-learning-based controllers. Although IPC (middle column) exhibits more distinct steps in measurement distribution than p99 latency (left column), it is frequently less informative than other rarely used counters (3 examples in the right column), which exhibit the most defined steps in measurement distribution. Across all experiments, no single counter was found to be an optimal QoS proxy. In fact, the most informative counter observed for Nginx is among the worst observed for Resnet and only slightly more informative than direct QoS measurements. These behaviors can be intuitively explained by the varied sensitivity of each workload to specific resource contention behaviors. Combining these observations, given a relatively small measurement period, we can more accurately estimate whether a long-term QoS target will be achieved by combining multiple counters rather than just a single counter or direct QoS measurements. Nevertheless, learning these relationships online in a data-limited, public cloud environment (as opposed to the data-rich, offline environments of prior QoS prediction approaches \cite{cosco, PROMPT2022}) can introduce significant error that must be addressed for reliable operation.

\subsubsection{Uncertainty in System Response}
Learning effective resource allocation policies remains a challenge, even with ideal information sources, due to \textit{inertia} in system response. This inertia is caused by inherent dependencies upon stateful architectural elements, such as the last-level cache, and manifests as state-dependent degradation in workload performance \cite{Ubik2014}. As a result, even a theoretically appropriate resource allocation, effective immediately, cannot guarantee acceptable short-term QoS since the system itself cannot respond immediately. Waiting to observe these effects is, naturally, problematic since any inertia in subsequent corrective actions may further delay response. This phenomenon implies distinct benefits for fine-grained control (i.e., sub-second to millisecond-level) that gradually corrects sub-optimal resource allocation decisions and thereby overcomes performance degradation as it occurs. Learning optimal corrective actions, however, necessitates policy learning at similarly fine-grained intervals, which has generally remained infeasible for machine-learning-based controllers due to limitations in information sources described above. These insights are a guiding principle for our work that leverages fast QoS prediction in order to enable policy learning at a rate orders-of-magnitude faster than prior work. Regardless, applying this insight necessitates additional consideration for efficient online learning strategies, which we detail in Section \ref{sec:rapid}.

\begin{figure*}[t]
\centering
\includegraphics[width=0.99\textwidth]{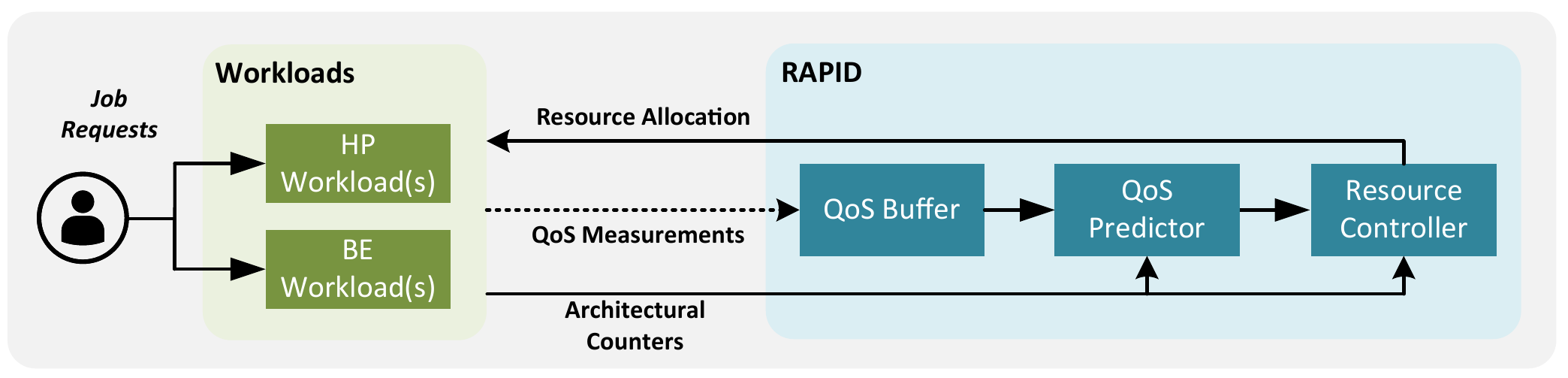}
\caption{RAPID framework for dynamic resource allocation.}
\label{fig:framework}
\end{figure*}


\section{RAPID} \label{sec:rapid}
In this section, we introduce RAPID, our machine-learning-based framework for fine-grained resource allocation. RAPID is proposed as a practical approach for fast, fully-online policy learning even in highly dynamic public cloud operating environments. Our novel approach is enabled by: efficiently integrating multiple information sources to achieve high QoS prediction accuracy with minimal sample requirements; decoupling policy learning from direct QoS measurements in order to reduce training time; and applying various domain-knowledge-inspired solutions for conventional problems in machine-learning-based resource allocation methods.


\subsection{Overview}
The high-level design for our proposed framework, RAPID, is illustrated in Figure \ref{fig:framework}. Framework execution begins upon receiving a new HP workload to be co-scheduled on the compute node. In general, we assume this HP workload to be previously unknown so do not have \textit{any} prior information about workload behaviors. RAPID therefore starts by sampling a small subset of resource allocation options in order to gather initial training data, given as pairs of architectural counters and QoS measurements. Once this initial sampling/training has completed, RAPID uses QoS predictions instead of QoS measurements to provide near-instantaneous feedback on resource allocation decisions. Our resource controller, based on deep reinforcement learning, takes advantage of this near-instantaneous feedback in order to safely explore alternative resource allocation options and exploit transient fluctuations in workload demand. QoS predictor accuracy is then continuously improved by accumulating long-term measurement averages, when available, while the primary control loop executes at a much finer granularity, thereby accommodating highly dynamic operating environments.


\begin{table*}[!b]
    \centering
    \caption{Selected architectural events (features).}
    \small
    \label{tab:selected_features}
        \begin{tabular}{lll}
            \toprule
            &\multicolumn{1}{l}{\textbf{Feature}} & \multicolumn{1}{l}{\textbf{Description} \cite{IntelManual}} \\
            \midrule
            \multirow{9}{2cm}{\textit{Counters (HP)}} & inst\_retired.any & Counts retired instructions \\ 
             & cpu\_clk\_unhalted.thread & Counts cycles when the core is not halted \\ 
             & frontend\_retired.latency\_ge\_32 & Counts period when no uops delivered for $\geq$ 32 cycles \\ 
             & uops\_executed.core\_cycles\_ge\_3 & Counts cycles when $\geq$ 3 uops are executed \\
             & rs\_events.empty\_cycles & Counts cycles when reservation station is empty \\
             & mem\_load\_retired.l2\_miss & Counts retired load instructions that miss L2 cache \\
             & cycle\_activity.stalls\_mem\_any & Counts cycles when execution pauses waiting for a load \\
             & offcore\_requests.all\_requests & Counts off-core memory transactions \\ 
             & offcore\_requests.demand\_data\_rd & Counts off-core memory transactions for demand reads \\
            \midrule \multirow{4}{2cm}{\textit{Counters (BE)}} & inst\_retired.any & Counts retired instructions \\
             & cpu\_clk\_unhalted.thread & Counts cycles when the core is not halted \\ 
             & l2\_rqsts.all\_demand\_miss & Counts demand requests that miss L2 cache \\
             & l2\_rqsts.swpf\_miss & Counts prefetch requests that miss L2 cache \\
             \bottomrule
		\end{tabular}
\end{table*}

\subsection{QoS Prediction} \label{sec:qos_prediction}
The fast QoS prediction in RAPID carefully balances data requirements and predictor accuracy through the use of multiple information sources, an efficient subset of resources and high-impact configurations, and a novel approach for predictor bias reduction.

\subsubsection{Model}
QoS prediction is implemented in RAPID as a regression task that maps short-term architectural counter measurements to long-term, per-workload QoS. Formally, QoS for HP workload $W_{w}^{HP}$ at time $t$ is predicted using a vector of architectural counter measurements $\mathbf{c}_{w,t}^{HP}$ and function approximator $f$, such that
\begin{equation}
P_{t}^{\,pred\,}(W_{w}^{HP}) = f(\mathbf{c}_{w,t}^{HP}).
\end{equation}
Recent work suggests that this mapping between architectural counters and QoS, although roughly linear \cite{cpi2_2013, LIBRA2021}, can be more accurately approximated by non-linear models \cite{Twig2020, PROMPT2022}. Nevertheless, overfitting becomes a distinct concern for non-linear models in our application due to limited training samples. We balance these concerns using a relatively low-complexity approach based on support vector regression with a radial basis function. This model, although non-linear, can be made to learn relatively smooth decision boundaries and, in testing, provided consistently lower prediction error than linear models, even when trained on just 20 samples. We further minimize predictor complexity by recognizing that current resource contention introduced by other workloads can be indirectly observed through architectural counters for each individual workload. In other words, predicting QoS for workload $W_{w}^{HP}$ should be possible even if the only input is counters from workload $W_{w}^{HP}$. Finally, RAPID uses only a small subset of available counters, listed as ``Counters (HP)'' in Table \ref{tab:selected_features}, which are selected by characterizing several representative cloud workloads following methodology in Penney et al. \cite{PROMPT2022}. Counters are aggregated across HP workload cores at the desired interval, log-transformed to accommodate order-of-magnitude differences in values, and then roughly normalized using interquartile range estimates.

\subsubsection{Reducing Training Data} 
RAPID implements several techniques to reduce training data requirements and overall training time. First, RAPID restricts sampling/allocation to an efficient subset of resources that have the highest impact on workload performance and power. This subset includes memory bandwidth, which enables RAPID to restrict the offcore request rate and practically eliminate contention for shared resources (e.g., last-level cache, memory, and network interfaces), and power, which RAPID uses to boost core frequency for compute-bound workloads.\footnote{This subset is selected purely for learning efficiency in common operating environments; other resources could be allocated with minor changes to controller action specification (Section \ref{sec:action_spec}).} Second, RAPID uses relatively coarse-grained resource allocation steps (e.g., adjust core frequency in 400 MHz steps rather than 100 MHz steps). Even with this approach, RAPID can still mimic fine-grained resource allocation behaviors by alternating between several steps. Finally, RAPID avoids uniform random sampling when gathering QoS prediction data due to the inherent bias that can occur with limited samples. Instead, RAPID leverages general domain knowledge in order to focus sampling on resource allocations that are most likely to affect resource contention and workload performance.

\subsubsection{Reducing Prediction Error}
Following QoS predictor training, the only \textit{required} measurements for framework operation are simple architectural counter measurements. Initial predictions, however, are likely to be biased towards workload behaviors observed during initial sampling. RAPID addresses this problem with a two-step solution that, first, estimates bias by comparing the distributions of past QoS measurements and predictions and, second, applies this estimate to compensate for bias in new predictions.

Past QoS measurements, although unreliable estimates for current QoS, can still provide useful insight into the range of workload performance levels that have actually occurred. Similarly, past QoS predictions reflect the range of workload performance levels that were expected to occur. As such, the difference between these two distributions can be used as an indicator for prediction bias. RAPID enables this comparison by maintaining a buffer with the previous $b$ seconds of QoS measurements and QoS predictions. A baseline estimate for workload performance, resulting from acceptable resource allocation, can be derived from lower percentile QoS measurements. We use the 10th percentile, denoted as $P_{[t-b\,,\,t]}^{\,meas\,,\,10\,}(W_{w}^{HP})$. This estimate is robust to measurement delays since average HP workload demand should remain relatively stable within $b$ seconds, provided $b$ is relatively small (e.g., 5 seconds). Conversely, estimates for degraded workload performance, resulting from poor resource allocations, can be derived from upper percentile values. We use the 90th percentile, denoted as $P_{[t-b\,,\,t]}^{\,pred\,,\,90\,}(W_{w}^{HP})$ for predictions and $P_{[t-b\,,\,t]}^{\,meas\,,\,90\,}(W_{w}^{HP})$ for measurements. Finally, bias for new QoS predictions $P_{t}^{\,pred\,}(W_{w}^{HP})$ at timestep $t$ can be corrected as shown in Equation \ref{eq:qos_correction}.
\begin{equation}
\label{eq:qos_correction}
\hat{P}_{t}^{\,pred\,}(W_{w}^{HP}) = P_{[t-b\,,\,t]}^{\,meas\,,\,10\,}(W_{w}^{HP}) + \dfrac{P_{t}^{\,pred\,}(W_{w}^{HP}) - P_{[t-b\,,\,t]}^{\,meas\,,\,10\,}(W_{w}^{HP})}{1 + \alpha\left(\dfrac{P_{[t-b\,,\,t]}^{\,pred\,,\,90\,}(W_{w}^{HP})}{P_{[t-b\,,\,t]}^{\,meas\,,\,90\,}(W_{w}^{HP})} - 1\right)}
\end{equation}
We introduce a gain factor $\alpha$ since, in a stable system, worst-case QoS measurements are expected to be lower than worst-case QoS predictions, given that predicted degradations in performance can be corrected before they actually occur. We empirically determined that $0.7\leq\alpha\leq0.9$ yielded desirable system behavior at a 0.2 second control interval.

\subsection{Resource Allocation Controller} \label{sec:controller}

RAPID enables fast, fully-online policy learning through efficient information re-use by a lightweight model architecture and action representation, effective use of QoS predictions as policy feedback, and domain-knowledge-inspired guidance for inertia-aware decisions and contention-aware exploration.

\begin{figure}[b]
\centering
\includegraphics[width=0.45\textwidth]{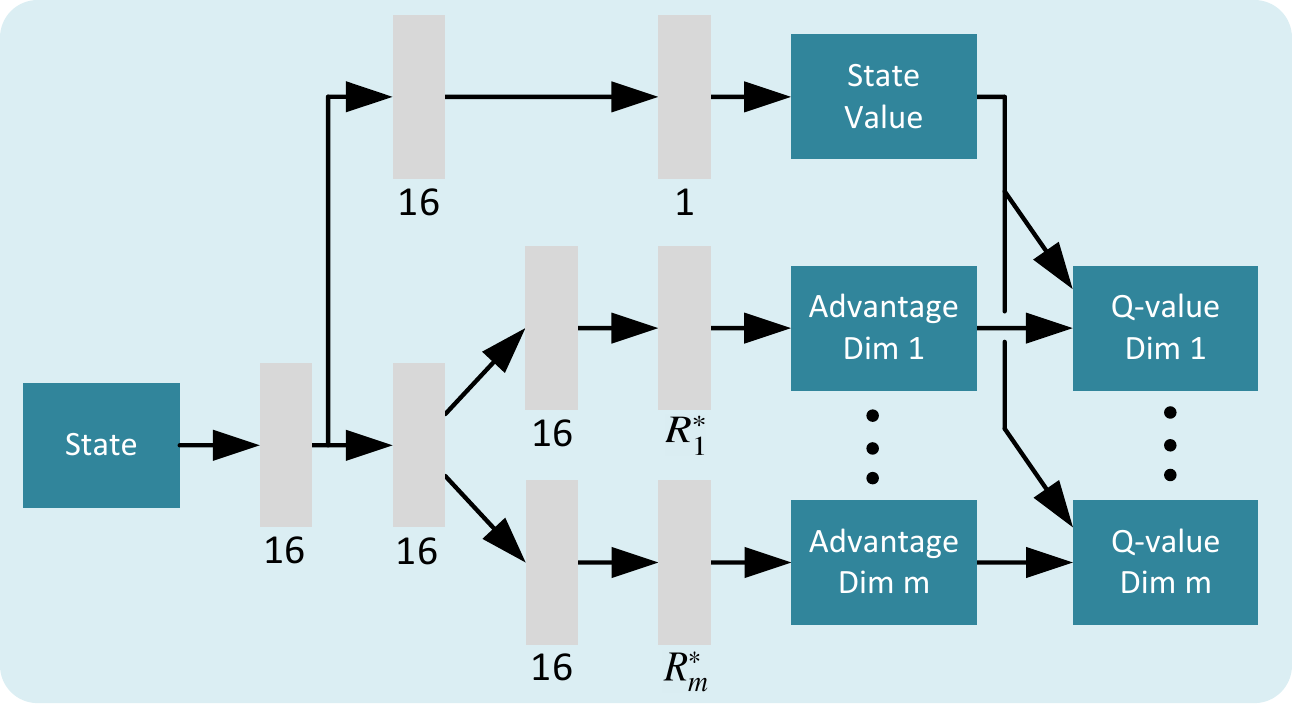}
\caption{Dynamic resource allocation controller architecture. Gray boxes represent fully connected layers, with output dimensionality listed below each box.}
\label{fig:rapid_bdq}
\end{figure}

\subsubsection{Model}
The resource controller in RAPID is implemented based on the action-branching architecture of Takavoli et al. and their variant of Dueling Double DQN, referred to as Branching Dueling Q-Network (BDQ) \cite{BDQ2019}. Our controller architecture, depicted in Figure \ref{fig:rapid_bdq}, uses a separate branch to approximate the advantage function for each action dimension, along with a unified state value predictor, in order to approximate the Q-value for each action dimension. Actions are then selected based on the configuration with the highest Q-value in each dimension. Splitting these action dimensions allows the number of networks outputs to grow linearly, rather than combinatorially, with respect to the action space, thus simplifying control complexity when allocating multiple resources. We calculate temporal-difference (TD) targets and losses following recommendations in Takavoli et al. \cite{BDQ2019}, with TD targets determined as the mean over all action dimensions and losses determined as the mean square error over all action dimensions. All hidden layers have 16 nodes, which we found to provide a good balance between model capacity and overall model stability given limited training data.

\subsubsection{State Specification}
State for the RAPID controller (Equation \ref{eq:state_spec}) consists of three elements. First, RAPID re-uses the HP workload architectural counters from QoS prediction $\mathbf{c}_{w,t}^{HP}$ as its primary information source to learn the effects of resource allocation on workload execution. Second, RAPID uses several BE workload counters, listed as ``Counters (BE)'' in Table \ref{tab:selected_features}, to calculate a normalized rate $\lambda_{t}^{BE}$ at which those BE workloads access shared resources during the execution. This normalized rate, given by Equation \ref{eq:weighted_l2_miss_ratio} (``cpu\_clk\_unhalted.thread'' abbreviated as $C$ and ``inst\_retired.any'' abbreviated as $I$), provides a nearly resource-independent estimate for the maximum \textit{potential} contention that could be caused by BE workloads. Finally, RAPID includes the current QoS, either predicted or measured, to provide the controller a direct gauge for its operating margin. All state information is log-transformed and normalized as described earlier for the QoS predictor.

\begin{equation}
\label{eq:state_spec}
\mathbf{S} = \{\ \mathbf{c}_{w,t}^{HP} \ , \ \lambda_{t}^{BE} \ , \ \hat{P}_{t}(W_{w}^{HP})\ \}
\end{equation}

\begin{equation}
\label{eq:weighted_l2_miss_ratio}
\lambda_{t}^{BE} = C * \dfrac{(l2\_rqsts.all\_demand\_miss + l2\_rqsts.swpf\_miss)}{I}
\end{equation}

\subsubsection{Action Specification} \label{sec:action_spec}
The dynamic nature of public clouds demands complete flexibility in terms of controller specification. We therefore cannot assume a fixed set of workloads as is done in many prior works (e.g., Twig \cite{Twig2020}). Instead, we specify actions for the HP workload only and then use those actions to indirectly select the resource allocation for BE workloads. Specification for each resource is as follows:
\begin{itemize}
    \item Memory bandwidth (MBW) actions ($a_{MBW}$) for HP workloads are based on the relative degree of \textit{restriction}, with possible values ranging from 0-90\%. These actions are then used to select the BE memory bandwidth restriction, specified as $100\%-a_{MBW}$. Although our test platform allows steps of 10\%, we instead select six configurations --- 0\%, 40\%, 60\%, 70\%, 80\%, and 90\% --- to reduce action space dimensionality as discussed earlier.
    \item Core frequency (CF) actions ($a_{CF}$) require consideration for system power limits, which impose frequency limits based on the number of cores. Limiting HP CF is therefore necessary to maximize BE CF. As such, we adopt a more complex overlap with six steps that translate $a_{CF}$ into $\{2400,3000,3300,3300,3300,3300\}$ for HP CF (in MHz) and $\{3300,3000,2400,2000,1400,800\}$ for BE CF.
\end{itemize}
The overall action space can then be represented as shown in Equation \ref{eq:action_space}.
\begin{equation}
\label{eq:action_space}
a_t = \{\{a_{MBW} \in \{0,...,5\}\}\ \times\  \{a_{CF} \in \{0,...,5\}\}\}
\end{equation}

\subsubsection{Reward Specification}
Rewards are specified to encourage resource allocation decisions that avoid QoS violations while maximizing resource allocation for BE workloads. Specifically, negative rewards (Equation \ref{eq:r_negative}) are given for actions leading to QoS violations, either during initial sampling when using QoS measurements or during regular operating when using QoS predictions (Equation \ref{eq:qos_ratio}). We then clip this penalty at $\beta=2.5$ to reduce noise in state-value/advantage estimates based on the intuition that all severe violations (e.g., more than 2.5x the target) are expected to require identical (maximal) throttling to quickly re-establish an appropriate QoS.
\begin{equation}
\label{eq:r_negative}
r_{w\,,\,t}^{-} = -\,\min\left(\,P_{t}^{\,ratio\,}(W_{w}^{HP})\:,\:\beta\,\right)
\end{equation}
\begin{equation}
\label{eq:qos_ratio}
P_{t}^{\,ratio\,}(W_{w}^{HP}) =\left\{
    \begin{array}{ll}
    \dfrac{P_{t}^{\,meas\,}(W_{w}^{HP})}{Q_{w}^{HP}}\,, & \mbox{if initial sampling}.\\[12pt]
    \dfrac{P_{t}^{\,pred\,}(W_{w}^{HP})}{Q_{w}^{HP}}\,, & \mbox{if regular operation}.\\
    \end{array}
  \right.
\end{equation}
Positive rewards (Equation \ref{eq:r_positive}) are given when no QoS violation is detected/predicted and are based on two components: resource allocation for BE workloads and QoS slack. Higher rewards are given for more aggressive BE resource allocations, among which individual resources can be prioritized using $\gamma$ when prior knowledge of \textit{BE} workloads is available, such as when a new HP workload is scheduled alongside a BE that had already been on a compute node for an extended duration. We empirically determine $\gamma=0.33$ to be an appropriate static value and leave dynamic tuning for future work. The final term is then used to encourage appropriate caution when operating near the QoS target (within 50\%) and generally favor resource allocation decisions with sufficient QoS slack to accommodate inertia in system response.
\begin{equation}
\label{eq:r_positive}
r_{w\,,\,t}^{+} = \gamma R_{CF\,,\,t}^{BE}\ +\ (1-\gamma) R_{MBW\,,\,t}^{BE}\ -\ \max\left(P_{t}^{\,ratio\,}(W_{w}^{HP})-0.5, 0\right)
\end{equation}

\subsubsection{Exploration}
Online learning renders exploration a potentially dangerous task. Randomly chosen resource allocations can cause severe QoS violations and also delay further exploration since BE workloads must be strictly throttled while processing backlogged requests. RAPID addresses this issue by selecting actions based on a discrete probability distribution around the current greedy action. Given exploration probability $\epsilon$ and a greedy action index $n$, the offset from the greedy action is given as
\begin{equation}
\label{eq:exploration}
o_{t} = \{-\dfrac{\epsilon}{2n},\,...\,,-\dfrac{\epsilon}{2n},1-\epsilon,\dfrac{\epsilon}{2}\}
\end{equation}
This distribution strictly clips the maximum offset to be +1 to account for the presence of QoS cliffs, in which a single resource allocation step can cause orders-of-magnitude degradation in QoS \cite{PROMPT2022}. Adopting this strict clipping also allows RAPID to maintain a high exploration probability throughout execution, thus supporting life-long exploration.

\section{Methodology}

\begin{table*}[b]
    \centering
    \caption{Characteristics summary for evaluated methods.}
    \small
    \label{tab:qualitative_summary}
        \begin{tabular}{lccccc}
            \toprule
            \multirow{3}{*}{\textbf{Method}} & \multicolumn{5}{c}{\textbf{Characteristic Category}} \\
            \cmidrule(lr){2-6}
             & \multirow{2}{2.2cm}{\centering\textbf{Online Policy Adaptation}} & \multirow{2}{2.2cm}{\centering\textbf{Time to\\Stable Policy}} & \multirow{2}{2.2cm}{\centering\textbf{Multi-Resource Allocation}} & \multirow{2}{2.2cm}{\centering\textbf{Multi-Objective Optimization}} & \multirow{2}{2.2cm}{\centering\textbf{Control Interval}} \\
             & & & & \\ \midrule
            \textbf{RL} & \cmark & Hours & \cmark & \cmark & Long \\
            \textbf{PID} & \xmark & Minutes & \xmark & \xmark & Short \\
            \textbf{RAPID} & \cmark & Minutes & \cmark & \cmark & Short \\
            \bottomrule
		\end{tabular}
\end{table*}

\subsection{Platform}
Experiments were conducted on a dual-socket server with Intel\textsuperscript{\textregistered} Xeon\textsuperscript{\textregistered} Gold 6336Y CPUs. All workloads were pinned to the first socket while traffic generators, representing user requests, were pinned to the second socket. Workloads were pinned to distinct subsets of cores on the first socket in order to prevent interference in the private cache, as is done in most prior work. Hyper-threading was enabled as it increases the maximum sustainable request rate for some workloads. Memory bandwidth was restricted using Intel\textsuperscript{\textregistered} Resource Director Technology \cite{RDT} while core frequency was managed using Intel\textsuperscript{\textregistered} Speed Select Technology \cite{SST}.

\subsection{Workloads} 
We selected a diverse set of workloads for experiments based on their prominence in cloud computing and general representativeness. HP workloads include:
\begin{itemize}
    \item Image Classification from the MLPerf Inference benchmark suite \cite{reddi2019mlperf}. This workload runs image classification with a ResNet-50 model \cite{Resnet} on the ImageNet dataset \cite{imagenet}.
    \item Recommendation from the MLPerf Inference benchmark suite. This workload uses a deep learning recommendation model (DLRM) on a synthetic Criteo Terabyte dataset \cite{criteo_terabyte}.
    \item Nginx web server \cite{nginx}. Nginx is the most widely-used web server and the preferred solution for the internet's most demanding service providers, including Netflix and Facebook \cite{nginx_popularity}. User requests are simulated using the wrk2 traffic generator \cite{wrk2}.
    \item Redis \cite{redis}. Redis is the most widely used key-value store engine, providing high-performance in-memory caching for a wide variety of data types \cite{redis_popularity}. User requests are simulated using the Locust traffic generator \cite{locust}.
\end{itemize}
BE workloads are selected from the SPEC CPU2017 benchmark suite \cite{SPEC} based on recommendations by Limaye and Adegbija \cite{limaye2018workload}. Selected workloads include 502.gcc, 507.cactuBSSN, 511.povray, 519.lbm, 520.omnetpp, 531.deepsjeng, 538.imagick, 548.exchange2, and 549.fotonik3d. These workloads span many domains including code compiling, video compression, fluid dynamics, atmospheric modeling, and more. 

We test two operating environments: 1) static HP load paired with dynamic BE load and 2) dynamic HP load paired with dynamic BE load. Appropriate request-per-second (RPS) rates for HP workloads were empirically determined as the rate at which no BE throttling is required (dynamic minimum); the rate at which strict BE throttling is required during periods of high BE demand (dynamic maximum); and the rate at which moderate BE throttling is required during periods of high BE demand (static). On our test platform, this corresponds to static/dynamic RPS rates of 250/100-350 for Image Classification, 250/100-350 for Recommendation, 400K/200K-500K for Nginx, and 65K/55K-110K for Redis. MLPerf workloads use the built-in traffic generator. For Nginx, we configure the wrk2 traffic generator to simulate 2K unique users accessing an HTML webpage. For Redis, we configure Locust to read/write 2K unique key/value pairs in batches of 64 with values of size 4KB. Dynamic BE job arrivals/completions are simulated by cycling the number of active threads, from a minimum of 3 to a maximum of 12, approximately every minute.


\subsection{Methods for Comparison}
We compare our proposed framework, RAPID, against the prior state-of-the-art methods in reinforcement learning (RL) \cite{Twig2020, PROMPT2022} and control theory (PID) \cite{LIBRA2021}. We summarize method characteristics in Table \ref{tab:qualitative_summary} and provide further detail below.

\textbf{RL:} This reinforcement-learning-based method uses a BDQ architecture for efficient learning with multiple action dimensions. This method, as originally proposed, supported co-scheduling with HP workloads only and took hours to train, so could not be used in public cloud environments. We therefore augment RL with \textit{all} controller-based optimizations used in RAPID, including model architecture and state/action/reward specification, along with all other optimizations proposed for Twig+ \cite{PROMPT2022}. As such, behavior differences between RL and RAPID can be attributed entirely to learning efficiency improvements proposed in this work. Initial training samples for RL are drawn from a uniform distribution, with ablation in Section \ref{sec:ablation}. Policy decisions/updates for RL are based on QoS measurements at a five second interval, which was determined to appropriately balance resource allocation delay and feedback consistency (Section \ref{sec:challenges}). 

\textbf{PID:} This control-theoretic approach uses a proportional-integral-derivative (PID) controller that provides consistent behavior with low overhead. Controller input is IPC and controller output is the memory bandwidth action \cite{LIBRA2021}. Since PID cannot control multiple resources, we set a safe, static allocation that maximizes HP core frequency. Memory bandwidth decisions are made by comparing the input IPC against a target IPC value determined during an initial sampling period. We find that eight samples (i.e., memory bandwidth configurations) at a five second measurement interval provides appropriate granularity when determining a target IPC, thus 40 seconds total for initial sampling. 
This static IPC threshold exemplifies a general limitation of conventional control methods as it is only guaranteed to remain valid under relatively static workload demands (i.e., similar to those observed during initial sampling) and may become sub-optimal in dynamic operating environments (Section \ref{sec:allocation_effectiveness}). Following initial sampling, PID uses a control interval of 200 milliseconds, below which testbench overhead (e.g., system calls to update resource allocation and gather architectural counters) becomes non-negligible.

\textbf{RAPID:} Our proposed framework first samples a small number of resource allocation options (e.g., 20), guided by domain knowledge, at a five second interval (i.e., 100 seconds total). These samples are used as initial training data for the QoS predictor and are also saved as experiences for replay. Afterwards, the sampling/action interval can be greatly reduced since RAPID uses QoS predictions to further guide policy learning. For fair evaluation, RAPID uses an identical control interval to PID (200 milliseconds) following initial sampling.

\subsection{Training Configuration}
\textbf{Reinforcement Learning:} We use the Adabelief optimizer \cite{zhuang2020adabelief} with $\beta$=(0.9, 0.999) and $\epsilon$\,=\,1E-8. Training begins at step 25 with target network updates every 40 steps, a batch size of 64, and a relatively high learning rate of 1E-2, paired with an aggressive gradient clipping of 0.5, to accelerate training \cite{zhang2019gradient}. The discount factor ($\gamma$) is set to 0.8 since poor resource allocation actions can generally be corrected within a few subsequent actions. A weight decay of 1E-3 and dropout of 0.1 are used to discourage overfitting to any particular workload demand \cite{generalization_dqn_2018}.

\textbf{QoS Prediction:} Regularization parameter $C=5$ is determined using grid-search in the range [0.1, 50.0]. Penalty parameter $\epsilon=0.05$ is set to penalize differences in QoS measurements of more than 5\% in log scale (ln(1.05)$\approx$0.05). Tolerance for stopping is set to 0.05. All other settings use the defaults in the scikit-learn implementation for support vector regression \cite{scikit-learn}.

\subsection{Metrics} \label{sec:metrics}
Recent work \cite{Twig2020, PROMPT2022} has compared HP workload QoS within the context of two metrics: 1) QoS guarantee, which considers how often QoS violations occur, and 2) QoS tardiness, which considers the average severity of those QoS violations. Neither metric, in isolation, provides sufficient information to determine the effective impact on end users and, even when reported together, may not be easily interpreted, such as when one controller incurs frequent but minor violations while another incurs infrequent but severe violations. We therefore introduce weighted QoS violation percent as a single, holistic metric. As shown in Equation \ref{eq:weighted_qos_violation}, this metric is calculated by averaging over all measurement intervals, with each interval either meeting the QoS goal (weight of 0\%) or not meeting the QoS goal (weight given as the percent increase in measured QoS over the target QoS).
\begin{equation}
\label{eq:weighted_qos_violation}
Q_{w\,,\,weighted} = 100*\dfrac{1}{T}\sum_{t\in T} \max\left(\dfrac{P_{t}^{\,meas\,}(W_{w})}{Q_{w}^{HP}}, 1\right)-1
\end{equation}

\begin{figure*}[t]
\centering
\includegraphics[width=1.0\textwidth]{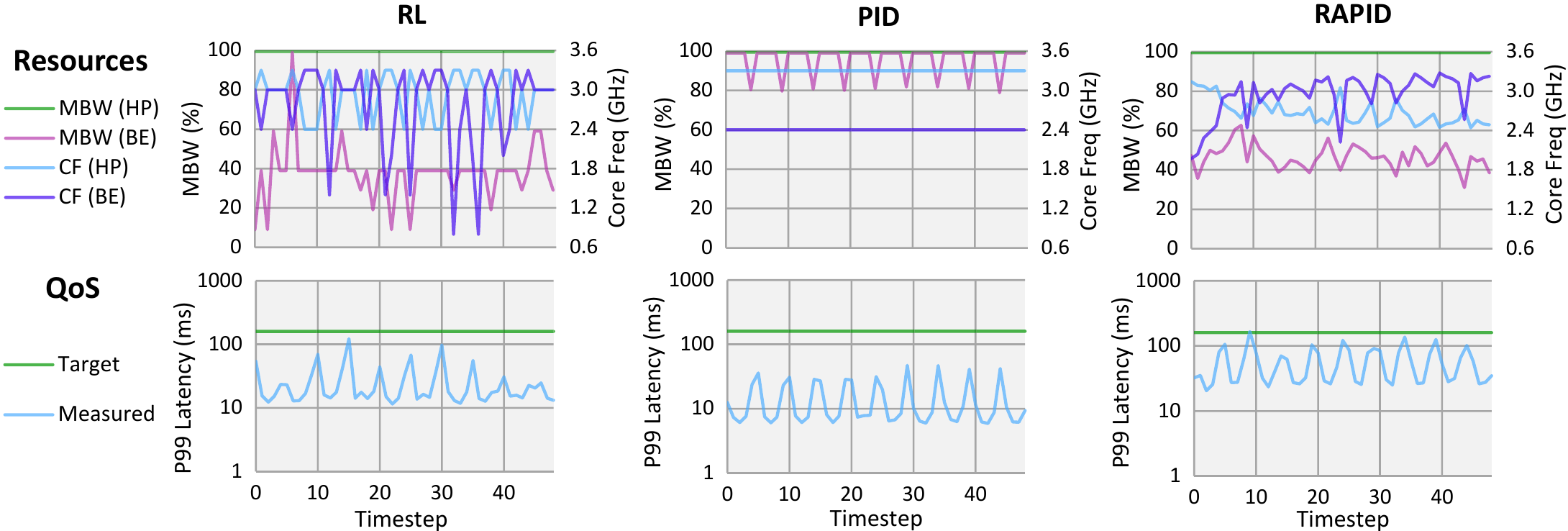}
\caption{Behavior comparison when co-scheduling Redis (HP) with 519.lbm (BE). All graphs use five-second averages to illustrate an identical action interval. RAPID learns consistent resource allocation policies that greatly reduce QoS slack (i.e., wasted resources).}
\label{fig:behavior_analysis}
\end{figure*}

\section{Evaluation} \label{sec:evaluation}

\begin{table*}[b]
    \centering
    \caption{Comparison of QoS and workload performance. Results show averages over all BE workloads. Experiments with static HP demand were repeated 3 times and confidence intervals were calculated across HP workloads. BE workload performance is normalized to that of RL. Abbreviations: Image Classification = IC; Recommendation = Rec.}
    \small
    \label{tab:empirical_comparison}
        \begin{tabular}{llrrrrr|rrrrr}
            \toprule
            \multirow{2}{*}{\textbf{Criteria}} & \multirow{2}{*}{\textbf{Method}} &
            \multicolumn{5}{c}{\textbf{Static HP Demand}} &
            \multicolumn{5}{c}{\textbf{Dynamic HP Demand}} \\
            \cmidrule(lr){3-7} \cmidrule(lr){8-12}
             & & \multicolumn{1}{c}{IC} & \multicolumn{1}{c}{Rec} & \multicolumn{1}{c}{Nginx} & \multicolumn{1}{c}{Redis} & \multicolumn{1}{c}{\textbf{Avg}} & \multicolumn{1}{c}{IC} & \multicolumn{1}{c}{Rec} & \multicolumn{1}{c}{Nginx} & \multicolumn{1}{c}{Redis} & \multicolumn{1}{c}{\textbf{Avg}}  \\
            \midrule
            \multirow{3}{2.3cm}{BE Performance (\textit{Higher=Better})} & RL & 1.00 & 1.00 & 1.00 & 1.00 & 1.00 $\pm$ 0.04 & 1.00 & 1.00 & 1.00 & 1.00 & 1.00 \\
             & PID & 1.14 & 1.21 & 1.17 & 1.26 & 1.19 $\pm$ 0.01 & 1.00 & 1.11 & 0.93 & 1.04 & 1.02 \\
             & \textbf{RAPID} & \textbf{1.33} & \textbf{1.41} & \textbf{1.44} & \textbf{1.55} & \textbf{1.43 $\pm$ 0.01} & \textbf{1.15} & \textbf{1.14} & \textbf{1.24} & \textbf{1.33} & \textbf{1.21} \\
            \cmidrule(lr){1-12}
            \multirow{3}{2.3cm}{Weighted QoS Violation \% (\textit{Lower=Better})} & RL & 0.097 & 0.036 & 0.000 & 0.000 & 0.033 $\pm$ 0.025 & 0.097 & 0.134 & 1.392 & 1.319 & 0.740 \\
             & PID & 0.000 & 0.000 & 0.000 & 0.000 & 0.000 $\pm$ 0.000 & 0.099 & 0.662 & 0.029 & 0.316 & 0.277 \\
             & \textbf{RAPID} & \textbf{0.000} & \textbf{0.000} & \textbf{0.000} & \textbf{0.001} & \textbf{0.000 $\pm$ 0.000} & \textbf{0.000} & \textbf{0.000} & \textbf{0.101} & \textbf{0.022} & \textbf{0.031} \\
            \bottomrule
		\end{tabular}
\end{table*}

\subsection{QoS Prediction}
We first examine QoS prediction model accuracy. Direct comparison with prior work is not possible as, to our knowledge, all related work using QoS prediction assumes substantial offline profiling and/or prior knowledge of workload behaviors. Instead, we compare our SVM-based approach with a representative boosted tree (BT) predictor \cite{PROMPT2022} and neural network (NN) predictor \cite{cosco}. For this, we focus on a scenario with 20 randomly selected samples, including 15 with acceptable QoS and 5 with unacceptable QoS, simulating expected behavior immediately following the initial sampling period. We report results over 25 independent trials when predicting p99 latency in milliseconds. Mean average error (MAE) for our SVM predictor (33) is substantially lower that of NN (45). BT has a similar MAE (32) but incurs approximately 1.4x higher false negatives than our SVM predictor. We further evaluate benefits when applying bias correction to our SVM predictor. Notably, applying bias correction was found to reduce QoS prediction error in all control tests, with an average reduction of 22.8\% for Redis and 10.1\% overall.

\subsection{Behavior Comparison}
Next, we examine differences in control behaviors resulting from sub-second resource allocation and online policy adaptation. As shown in Figure \ref{fig:behavior_analysis}, RL exhibits relatively erratic policy decisions and is unable to consistently minimize QoS slack (i.e., the difference between the measured and target QoS; larger slack indicates more wasted resources). In contrast, RAPID leverages its fast decision interval to gather much more training data and converges to a stable resource allocation policy within approximately 50 seconds (timestep 10 in Figure \ref{fig:behavior_analysis}). Furthermore, the policy learned by RAPID is shown to provide more consistent reduction in QoS slack. In general, RAPID is able to explore more aggressive resource allocation options since any degradations in workload performance can be corrected faster than with RL due to differences in their resource allocation interval. PID inherently provides high consistency in resource allocation decisions under most conditions yet, as shown, cannot appropriately reduce QoS slack in all scenarios. Further, there may exist edge-case scenarios in which QoS degrades significantly without measurably impacting IPC. In contrast, the additional architectural counters used by RAPID provide alternative perspectives into resource contention and, therefore, enable more robust behavior.

\subsection{Resource Allocation Effectiveness} \label{sec:allocation_effectiveness}
We compare the benefits of RAPID against RL and PID across a variety of co-scheduling scenarios including four prominent HP workloads and nine representative BE workloads. An ideal resource allocation policy should maximize BE workload performance with minimal impact on QoS. Results are given in Table \ref{tab:empirical_comparison}. First, we consider BE workload performance, measured as instructions-per-second. RL performs worst, both for static and dynamic HP demand, as it requires much more time to learn an appropriate resource allocation policy and must also be more conservative in accommodating inertia due to its longer decision interval. PID provides a 19\% improvement over RL in tests with static HP demand, but just 2\% improvement over RL in tests with dynamic HP demand since its IPC threshold, determined based on workload demand at the start of execution, can become inappropriate as operating conditions change. In contrast, RAPID offers the highest BE performance in all scenarios, with average improvements over RL of approximately 43\% for static and 21\% for dynamic HP demand, as RAPID both quickly and continuously adapts to changing conditions.

The second consideration is QoS, which we report as the weighted QoS violation percent, taking into account both violation frequency and severity (Section \ref{sec:metrics}). This number should, ideally, be as close as possible to zero since even a small number of QoS violations can lead to substantial penalties, potentially millions of dollars, as well as higher operational costs. Even with static HP demand, RL incurs a significant number of QoS violations near the start of training, although most are only slightly above the QoS target, resulting in a modest QoS degradation of 0.03\%. These violations, however, become more significant in tests with dynamic HP demand, in which inappropriate resource allocation decisions in previously unseen states can cause substantial violation (0.74\%) in average QoS. Interestingly, PID maintains near-ideal QoS in some scenarios yet significantly degrades QoS in others, likely due to differences in sensitivity to the initial IPC threshold, as individual workloads may exhibit larger IPC changes due to load than others. Finally, RAPID is shown to provide the strongest QoS guarantees, on average, as any inappropriate policy decisions can be quickly detected and corrected. Specifically, with dynamic HP demand, RAPID reduces QoS degradation by 95.8\% compared with RL and 88.9\% compared with PID (9.0x improvement). Overall, RAPID achieves strong QoS guarantees while greatly improving BE workload performance.

\subsection{Overhead} \label{sec:overhead}
RAPID is designed to be lightweight and supports millisecond-level resource allocation. Specifically, we measured execution time for inference \textit{on a single core} to be roughly 0.2 ms for the QoS predictor and 1.5 ms for the resource controller, corresponding to an overhead of just 0.85\%. This overhead is easily outweighed by gains in BE performance and could be further reduced simply by thresholding against currently predicted QoS slack (i.e., only execute the controller when re-allocation is likely to be beneficial). Overhead for QoS predictor training (5ms) and resource controller training (10ms) can generally be ignored since they are not on the critical path for resource allocation decisions and can be performed infrequently following the initial learning period. As such, RAPID could feasibly operate at a 5-10 ms resource allocation interval in an embedded implementation without overhead from system calls.

\subsection{Ablation} \label{sec:ablation}
Several ablation tests were conducted to verify the impact of proposed components. First, we observed that augmenting RL with domain-knowledge guidance during initial sampling, rather than using uniform random samples, led to no statistically significant change in BE performance and only modest improvement in HP QoS (still $\approx$10x higher weighted QoS violations than RAPID). Intuitively, these initial samples can affect early policy learning, but quickly become obsolete as workload dynamics change over time. Second, we found that removing all counters except IPC from RAPID incurs a 5.9x increase in weighted QoS as lower quality guidance allows for overly aggressive resource allocation policies. Finally, removing bias correction from RAPID notably decreases BE performance (5\% average, 10\% on Redis) since there are more instances with underpredicted QoS slack during which RAPID unnecessarily throttles BE workloads.

\subsection{Discussion} \label{sec:discussion}

\subsubsection{Theoretical Analysis}
Precisely modeling workload performance is impractical in our targeted operating environment due to the inherent lack of knowledge about workload behaviors and resource demands. Specifically, it is impossible to guarantee that a certain resource allocation will never cause a QoS violation. We can, however, establish criteria under which RAPID guarantees less severe QoS violations. Here, we assume an identical resource allocation policy for RL and RAPID. We then define the severity of a QoS violation, starting at $t_1$ and ending at $t_2$ as shown in Equation \ref{eq:qos_severity}.
\begin{equation}
\label{eq:qos_severity}
S = \int_{t_1}^{t_2} Perf(W^{HP}(t)) - Q^{HP}\,dt 
\end{equation}
The faster control interval in RAPID reduces delay in responding to changing workload state and, consequently, allows demands to be satisfied faster, thus reducing $t_2 - t_1$. Responses made by RAPID will also be more appropriately proportional to resource demand since state, which includes QoS, would nearly immediately reflect degraded workload performance when using QoS predictions but lags when using QoS measurements. In other words, RAPID is expected to reduce both the duration and magnitude of QoS violations. These advantages also translate to higher BE performance since RAPID can respond more quickly and more appropriately to periods with low HP resource demand.

\subsubsection{Multiple Workloads and Multiple Compute Nodes}
Extending reinforcement-learning-based controllers to multiple workloads across multiple compute nodes is shown by related work to be a relatively straightforward task \cite{firm2020}. Specifically, a separate controller for each HP workload was found to outperform solutions with a single controller for all HP workloads. This approach is directly applicable to our work since each RAPID controller attempts to minimize the resources for its HP workload, thereby collectively maximizing resources available for all other workloads. Higher-level coordination would still be maintained by a cluster scheduler, which complements RAPID by mitigating concerns of long-term interference between HP workloads due to insufficient node resources.

\subsubsection{Other Control Methods}
Control methods based on Bayesian optimization have become popular in recent work due to their simplicity and reliability. Regardless, these methods generally assume a relatively static operating environment in which resource allocation configurations remain optimal for an extended period. These assumptions become problematic in more general operating environments such as that targeted by our work. In practice, accommodating the dynamic fluctuations in resource demand illustrated in Figure \ref{fig:behavior_analysis} would require near-continuous re-sampling or very conservative resource allocations. These disadvantages are highlighted in related work \cite{PROMPT2022} so we do not consider Bayesian optimization in our testing.

\section{Conclusion}

Workload co-scheduling, enabled by fine-grained resource allocation, has become a promising method for cloud service providers to meet growing demand without compromising customer requirements. Nevertheless, resource allocation approaches based on machine learning have remained largely impractical in highly dynamic operating environments such as the public cloud. Our proposed approach, RAPID, addresses these limitations by decoupling policy decisions from conventional feedback sources using a lightweight QoS prediction model. These predictions are shown to provide effective feedback for policy decisions at a rate orders of magnitude faster than is possible with QoS measurements, thereby enabling fully-online policy learning in highly dynamic operating environments. Evaluation confirms that RAPID can learn stable resource allocation policies in just minutes as compared with hours in prior work.
As a result, RAPID substantially improves both HP QoS and BE workload performance, which translate into higher customer satisfaction and reduced overall cost through more effective resource utilization.


\section*{Acknowledgments}
We thank Kamil Andrzejewski for assistance with PID controller implementation and Ripan Das for his assistance with workload setup. This work was supported, in part, by Intel Corporation's Academic Research funding.


\bibliographystyle{elsarticle-num} 
\bibliography{refs}





\end{document}